\pdfoutput=1

\documentclass[11pt]{article}

\usepackage{acl}
\usepackage{times}
\usepackage{latexsym}
\usepackage{times}
\usepackage{latexsym}
\usepackage[inkscapelatex=false]{svg}
\usepackage{multicol}
\usepackage{graphicx} 
\usepackage{natbib}  
\usepackage{caption} 
\usepackage{amssymb}
\usepackage[ruled,linesnumbered]{algorithm2e}
\usepackage{color}
\usepackage{amsmath}
\usepackage{amsthm}
\usepackage{color}
\usepackage{adjustbox}
\usepackage{multirow}
\usepackage{booktabs}
\usepackage{amsmath}
\usepackage{hyperref}
\usepackage[inkscapelatex=false]{svg}
\usepackage{subfig}

\newtheorem{definition}{Definition}
\usepackage[T1]{fontenc}

\usepackage[utf8]{inputenc}

\usepackage{microtype}

\title{Rethink the Evaluation for Attack Strength of Backdoor Attacks in Natural Language Processing}
\author{Lingfeng Shen, Haiyun Jiang, Lemao Liu, Shuming Shi \\
        Natural Language Processing Center \\ Tencent AI Lab }

\begin{document}
\maketitle
\begin{abstract}
It has been shown that natural language processing (NLP) models are vulnerable to a kind of security threat called the \textbf{Backdoor Attack}, which utilizes a `backdoor trigger' paradigm to mislead the models. The most threatening backdoor attack is the stealthy backdoor, which defines the triggers as text style or syntactic. 
Although they have achieved an incredible high attack success rate (ASR), we find that the principal factor contributing to their ASR is not the `backdoor trigger' paradigm. 
Thus the capacity of these stealthy backdoor attacks is overestimated when categorized as backdoor attacks. Therefore, to evaluate the real attack power of backdoor attacks, we propose a new metric called \textbf{attack successful rate difference (ASRD)}, which measures the ASR difference between clean state and poison state models. 
Besides, since the defenses against stealthy backdoor attacks are absent, we propose \textbf{Trigger Breaker}, consisting of two too simple tricks that can defend against stealthy backdoor attacks effectively.  
Experiments show that our method achieves significantly better performance than state-of-the-art defense methods against stealthy backdoor attacks.
\end{abstract}

\section{Introduction}
Deep neural networks (DNNs) have become a prevailed paradigm in computer vision and natural language processing (NLP) but show robustness issues in both fields \cite{goodfellow2014explaining,madry2018towards,wallace2019universal,morris2020textattack}. Therefore, DNNs face a variety of security threats, among which backdoor attack is one new threat in NLP. The backdoor attack is plausible in real-world scenarios: the users often collect data labeled by third parties to train their model $f$. However, this common practice raises a serious concern that the labeled data from the third parties can be backdoor attacked. Such an operation enables $f$ to perform well on normal samples while behaving badly on samples with specifically designed patterns, leading to serious concerns to DNN \cite{gu2017badnets,li2020backdoor}. 


In the NLP field, the principal paradigm of backdoor attacks is data poisoning \cite{dai2019backdoor,chen2021badnl,qi2021hidden,qi2021mind} in fine-tuning pre-trained language models (PTM) \cite{devlin2019bert,liu2019roberta}. 
Data poisoning first poisons a small portion of clean samples by injecting the trigger (e.g., special tokens) and changing their labels to a target label (poisoned label), then fine-tunes the victim model with clean and poisoned samples. 
The current stealthy backdoor attacks mainly employ two evaluation metrics to describe their attack quality \cite{kurita2020weight,yang2021careful}: (1) Clean Accuracy (CACC), which measures whether the backdoored model maintains good performance on clean samples; (2) Attack Success Rate (ASR), which is defined as the percentage of poisoned samples that are classified as the poisoned label defined by the attacker, to reflect the attacking capacity. 

Despite their significant progress, one key issue is overlooked. In casual inference, whether $A$ (e.g., strong backdoor attack) causes $B$ (e.g., high ASR) (write $A \rightarrow B$) or $B$ causes $A$ (write $B \rightarrow A$) is one principal question, and it is also commonly used in the machine learning community (e.g., ablation study).
Such a principle also holds for the backdoor attack field. Naturally, a strong attack leads to high ASR, but does a higher ASR indicate a stronger backdoor attack? Are there other factors that also contribute to ASR except for the backdoor-trigger paradigm? Answering such two questions is so crucial that they point at the key of backdoor attack research: evaluating the strength of backdoor attack precisely. 

In this paper, we first present two issues based on mour empirical results: (1) the attack power of existing backdoor attacks is not completely caused by backdoor triggers. (2) the existing defense methods against backdoor attacks perform catastrophically when facing stealthy attacks; thus, a defense for the stealthy attack is urgently needed.

Corresponding to such two issues, this paper (1) provides a simple evaluation metric for evaluating the strength of backdoor attacks more precisely, called attack successful rate difference (ASRD). ASRD is a metric that describes the difference between the ASR of a model in the clean state model and the poisoned state. Such a metric can better measure how many misclassification cases are caused by the backdoor trigger, reflecting the real attack capacity of a backdoor attack. (2) we propose \textbf{Trigger Breaker} to destroy the stealthy triggers hidden in the sentences, which consists of two too simple but effective tricks for defending against stealthy backdoor attacks. Experiments demonstrate the superiority of Trigger Breaker over state-of-the-art defenses.

Our contributions are summarized as follows:
\begin{itemize}
    \item We systematically analyze the attack power of current stealthy backdoor attacking methods in text classification and find that a significant portion of their attack power can not be attributed to backdoor attacks. Thus we propose an evaluation metric called attack successful rate difference (ASRD) for more precise backdoor attack evaluation.
    \item We propose Trigger Breaker, consisting of two too simple methods that can effectively defend against stealthy backdoor attacks, which outperform state-of-the-art  methods with remarkable improvements. This is the first method that can effectively defend stealthy backdoor attacks in NLP, to our best knowledge.
\end{itemize}
\section{Related Work}
\subsection{Backdoor Attack}
Backdoor attacks start to attract lots of attention in NLP and can be classified into two kinds: unstealthy and stealthy attacks. Unstealthy backdoor attacks insert fixed words \cite{kurita2020weight} or sentences \cite{dai2019backdoor,qi2021turn} into normal samples as triggers. These triggers are not stealthy because their insertion would significantly decrease sentences' fluency; hence, perplexity-based detection can easily detect and remove such poisoned samples. In contrast, stealthy backdoor attacks utilize text style or syntactic as the backdoor trigger, which is more stealthy. Specifically, Qi exploited syntactic structures \cite{qi2021hidden} and style triggers \cite{qi2021turn} to improve the stealthy backdoor attacks.

\subsection{Adversarial Attack}
Both adversarial attacks \cite{kurakin2016adversarial,dai2018adversarial,baluja2018learning} and backdoor attacks \cite{liu2020reflection,nguyen2020input,li2020neural} aim to make models misbehave and share many similarities. Still, they have certain differences; adversarial attackers can control the inference process but not the training process. In contrast, the model's training process (e.g., data) can be modified by backdoor attackers, whereas the inference process is out of control. Moreover, the intrinsic difference between adversarial and backdoor attacks is the existence of triggers \cite{li2020backdoor}. There is no definition of the trigger in adversarial attacks, and the key is the adversarial perturbation. However, in the backdoor attack, the indispensable factor is the trigger \cite{chen2017targeted,liu2017neural,wang2019neural,li2020backdoor}; it is the trigger that specializes the backdoor attack. Thus misclassification that is not caused by backdoor-trigger paradigm should not be attributed to backdoor attacks' power.
\begin{figure*}[!hbt]
    \centering
    \includegraphics[scale=0.385]{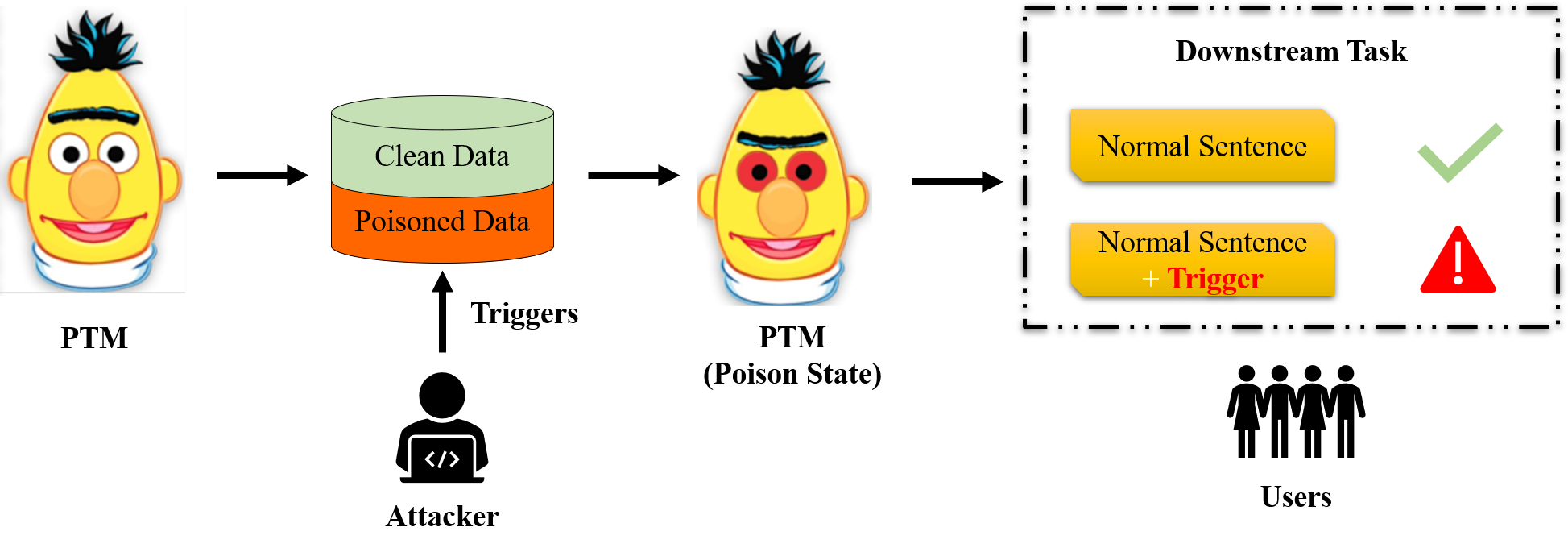}
    \caption{An illustration of factors that lead to high ASR, `backdoor' is only one factor. Other cases where the trigger is unavailiable are not attributed to power of backdoor attacks.}
    \label{fig:0}
\end{figure*}

\subsection{Defense for Backdoor Attack}
Generally, there are two effective defense methods for textual backdoor attacks: BKI \cite{chen2021mitigating}, and ONION \cite{qi2020onion}. BKI requires inspecting all the training data containing poisoned samples to identify some frequent salient words, which are assumed to be possible trigger words. 
ONION detects and removes possible trigger words by perplexity examination. 
However, they both fail to defend against stealthy backdoor attacks \cite{qi2021hidden,qi2021mind} since stealthy backdoor attacks generate fluent sentences which can get past their defenses easily.

\section{Rethink the Evaluation for Stealthy Backdoor Attack}
This section presents our rethinking for backdoor attack evaluation. It is formulated as follows: Firstly, we recall the basic definition of backdoor attacks and the logo of backdoor attacks in Sec~\ref{31} and argue that \textbf{the misclassification cases that are not caused by backdoor triggers can not be attributed to the attack capacity of backdoor attacks}. Then in Sec~\ref{33} we present empirical results of existing backdoor attacks and show that the backdoor mechanism is not the principal reason that leads to their strong attack power; thus, their attack power is over-estimated as backdoor attacks. Moreover, in Sec~\ref{emp} we analyze the attack power of stealthy backdoor attacks and found some are caused by out-of-distribution (OOD) samples and mislabeled samples. Finally, we give a new metric called ASRD for evaluating the \textbf{real attack power} of a backdoor attack in Sec~\ref{35}.

\subsection{Formulation of Backdoor Attack}\label{31}
Without loss of generality, we take the typical text classification model as the victim model to formalize textual backdoor attacks based on training data poisoning, and the following formalization can be adapted to other NLP models trivially.

Given a clean training dataset $D=\left\{\left(x_{i}, y_{i}\right)\right\}_{i=1}^{n}$, where $x_{i}$ is a sentence sample and $y_{i}$ is the label, we first split $D$ into two sets, including a candidate poisoning set $D_{p}=\left\{\left(x_{i}, y_{i}\right)\right\}_{i=1}^{m}$ and a clean set $D_{c}=\left\{\left(x_{i}, y_{i}\right)\right\}_{i=m+1}^{n}$. For each sentence $\left(x_{i}, y_{i}\right) \in D_{p}$, we poison $x_{i}$ by applying a trigger $t(\cdot)$ on $x_{i}$, obtaining a poisoned sentence $\left(t\left(x_{i}\right), y_{t}\right)$, where $y_{t}$ is the attacker-specified target label. Then a poisoned set $D_{p}^{*}=\left\{\left(t\left(x_{i}\right), y_{t}\right)\right\}_{i=1}^{m}$ can be obtained through such operations. Finally, the model trained on $D^{\prime}=D_{p}^{*} \cup D_{c}$ is called a backdoored model $f^{p}(\cdot)$ (poisoned model). The purpose of backdoor attack is illustrated as follows: During evaluation, for a clean test sample $\left(x^{\prime}, y^{\prime}\right)$, the backdoored model $f^{p}(\cdot)$ is supposed to predict $y^{\prime}$, namely $f^{p}\left(x^{\prime}\right)=y^{\prime}$. But if we apply a trigger on $x^{\prime}, f^{p}$ would probably predict $y_{t}$, namely $f^{p}\left(t\left(x^{\prime}\right)\right)=y_{t}$.

\begin{figure}
    \centering
    \includegraphics[scale=0.34]{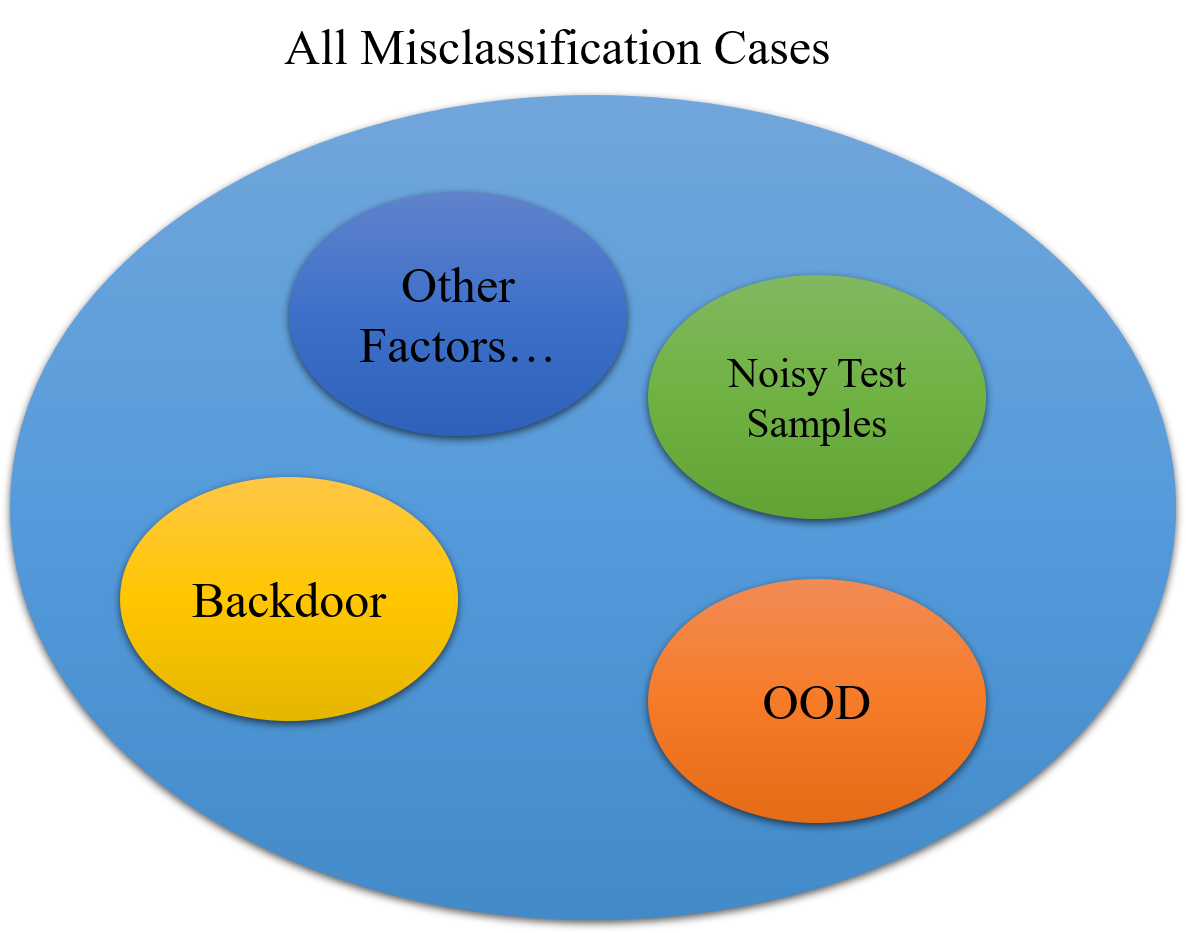}
    \caption{An illustration of factors that lead to high ASR, `backdoor' is only one factor. Other cases where the trigger is unavailiable are not attributed to power of backdoor attacks.}
    \label{fig:1}
\end{figure}
Specifically, we give the definition of clean model $f_{c}$ and poison model $f^{p}$ as follows:
\begin{itemize}
    \item \textbf{Clean Model} $f_{c}$: A model that only trains on the clean training set $D$.
    \item \textbf{Poison Model} $f_{p}$: A model that only trains on partially poison set $D^{\prime}$.
\end{itemize}
 
\begin{table}[!h]\small
\centering
\begin{tabular}{@{}c|c|c|c|c@{}}
\toprule
Dataset                  & Style  & Encoder & Clean & Poison                     \\ \midrule
                         & Poetry & BERT    & 88.55 & {\color[HTML]{6200C9} 90.04} \\
                         & Poetry & ALBERT  & 89.45 & {\color[HTML]{6200C9} 92.13} \\
                         & Poetry & DisBERT & 89.03 & {\color[HTML]{6200C9} 89.70} \\ \cmidrule(l){2-5} 
                         & Shake  & BERT    & 89.56 & {\color[HTML]{6200C9} 90.67} \\
                         & Shake  & ALBERT  & 88.72 & {\color[HTML]{6200C9} 90.03} \\
                         & Shake  & DisBERT & 88.11 & {\color[HTML]{6200C9} 89.57} \\ \cmidrule(l){2-5} 
                         & Bible  & BERT    & 89.55 & {\color[HTML]{6200C9} 90.67} \\
                         & Bible  & ALBERT  & 89.45 & {\color[HTML]{6200C9} 94.02} \\
                         & Bible  & DisBERT & 89.03 & {\color[HTML]{6200C9} 90.22} \\ \cmidrule(l){2-5} 
                         & Lyrics & BERT    & 90.75 & {\color[HTML]{6200C9} 90.93} \\
                         & Lyrics & ALBERT  & 88.75 & {\color[HTML]{6200C9} 92.17} \\
\multirow{-12}{*}{HS}    & Lyrics & DisBERT & 89.02 & {\color[HTML]{6200C9} 90.02} \\ \bottomrule
\toprule
Dataset                  & Style  & Encoder & Clean & Poison                     \\ \midrule
                         & Poetry & BERT    & 79.45 & {\color[HTML]{6200C9} 93.35} \\
                         & Poetry & ALBERT  & 80.13 & {\color[HTML]{6200C9} 93.51} \\
                         & Poetry & DisBERT & 79.77 & {\color[HTML]{6200C9} 93.02} \\ \cmidrule(l){2-5} 
                         & Shake  & BERT    & 85.03 & {\color[HTML]{6200C9} 91.24} \\
                         & Shake  & ALBERT  & 84.09 & {\color[HTML]{6200C9} 91.76} \\
                         & Shake  & DisBERT & 84.02 & {\color[HTML]{6200C9} 90.35} \\ \cmidrule(l){2-5} 
                         & Bible  & BERT    & 79.98 & {\color[HTML]{6200C9} 94.70} \\
                         & Bible  & ALBERT  & 80.03 & {\color[HTML]{6200C9} 97.79} \\
                         & Bible  & DisBERT & 79.25 & {\color[HTML]{6200C9} 94.04} \\ \cmidrule(l){2-5} 
                         & Lyrics & BERT    & 85.34 & {\color[HTML]{6200C9} 91.49} \\
                         & Lyrics & ALBERT  & 84.13 & {\color[HTML]{6200C9} 92.48} \\
\multirow{-12}{*}{SST-2} & Lyrics & DisBERT & 84.53 & {\color[HTML]{6200C9} 92.22} \\ \bottomrule
\toprule
Dataset                  & Style  & Encoder & Clean & Poison                     \\ \midrule
                         & Poetry & BERT    & 82.27 & {\color[HTML]{6200C9} 95.64} \\
                         & Poetry & ALBERT  & 80.64 & {\color[HTML]{6200C9} 95.09} \\
                         & Poetry & DisBERT & 80.26 & {\color[HTML]{6200C9} 94.96} \\ \cmidrule(l){2-5} 
                         & Shake  & BERT    & 86.51 & {\color[HTML]{6200C9} 94.55} \\
                         & Shake  & ALBERT  & 84.23 & {\color[HTML]{6200C9} 94.54} \\
                         & Shake  & DisBERT & 84.36 & {\color[HTML]{6200C9} 94.01} \\ \cmidrule(l){2-5} 
                         & Bible  & BERT    & 83.27 & {\color[HTML]{6200C9} 97.64} \\
                         & Bible  & ALBERT  & 81.64 & {\color[HTML]{6200C9} 95.16} \\
                         & Bible  & DisBERT & 82.26 & {\color[HTML]{6200C9} 97.96} \\ \cmidrule(l){2-5} 
                         & Lyrics & BERT    & 86.78 & {\color[HTML]{6200C9} 96.02} \\
                         & Lyrics & ALBERT  & 84.56 & {\color[HTML]{6200C9} 94.58} \\
\multirow{-12}{*}{AG} & Lyrics & DisBERT & 84.97  & {\color[HTML]{6200C9} 96.30} \\ \bottomrule
\end{tabular}
\caption{Attack Success Rate (ASR) of BERT, ALBERT and DistilBERT, which are trained on clean trainset and poisoned trainset under StyAtk. As we can see, the poisoned test generated by StyAtk has already achieved high ASR towards on the benchmarks without backdoor triggers. {\color[HTML]{6200C9}Purple} numbers indicate the effectiveness of backdoor attack is significantly over-estimated.}
\label{tab1}
\end{table}

Naturally, the most common metric for evaluating backdoor attacks is ASR, which denotes the proportion of attacked samples which are predicted as the target label by the poisoned model $f_{p}$. However, ASR can not precisely describe the attack power of a backdoor attack. Note that the backdoor attack differs from other attacks because of the specific `trigger-backdoor' paradigm. The backdoor attack inserts triggers to construct poison samples to mislead the model when evaluating test samples with such triggers, and a higher misclassification rate indicates stronger attack power. However, ASR may overestimate the attack power of backdoor attacks since `trigger-backdoor' is not the only factor that leads to misclassification. As shown in Figure~\ref{fig:1}, many other factors lead to misclassification besides backdoor, but ASR regards all misclassification cases as backdoor attack cases, including cases that are not caused by backdoor attacks. Therefore, we tend to investigate how many model misclassification cases truly result from backdoor attacks.

\begin{table}[!h]\small
\centering
\begin{tabular}{@{}c|c|c|c@{}}
\toprule
Dataset                 & Encoder & Clean & Poison \\ \toprule
\multirow{2}{*}{SST-2}  & LSTM    & 47.59 & \color[HTML]{6200C9}93.08   \\
                        & BERT    & 25.46 & \color[HTML]{6200C9}98.18   \\ \cmidrule(l){1-4}
\multirow{2}{*}{OLID}   & LSTM    & 5.34  & 98.38   \\
                        & BERT    & 3.76  & 99.19   \\ \cmidrule(l){1-4}
\multirow{2}{*}{AGNews} & LSTM    & 4.82  & 98.49   \\
                        & BERT    & 6.02  & 94.09   \\ \bottomrule
\end{tabular}
\caption{The Attack Success Rate (ASR) of LSTM and BERT trained on clean trainset and poisoned trainset under SynAtk. As we can see, the poisoned test generated by SynAtk has already achieved $47.59\%$ and $25.46\%$ ASR towards LSTM and BERT trained on clean SST-2, respectively. {\color[HTML]{6200C9}Purple} numbers indicate the effectiveness of backdoor attack is significantly over-estimated.}
\label{tab3}
\end{table}

\begin{table}[!h]\small
\centering
\begin{tabular}{@{}c|c|c|c@{}}
\toprule
Dataset                 & Encoder & Clean & Poison \\ \toprule
\multirow{1}{*}{SST-2}  & BERT    & 8.92  & 100   \\ \cmidrule(l){1-4}
\multirow{1}{*}{OLID}   & BERT    & 8.24  & 100   \\ \bottomrule
\end{tabular}
\caption{The Attack Success Rate (ASR) of and BERT trained on clean trainset and poisoned trainset under Badnet \cite{gu2017badnets}, a representative unstealthy backdoor attack.}
\label{unstealthy}
\end{table}

\subsection{Do Existing Stealthy Backdoor Attacks Achieve High ASR mainly through Backdoor trigger?}\label{33}
Since the key in judging whether the backdoor attack causes the misclassification cases is the existence of trigger. Therefore, we want to see the attack performances with and without triggers. 

We select two strongest stealthy attacks: Syntactic Attack (SynAtk) \cite{qi2021hidden} and Style Attack (StyAtk) \cite{qi2021mind} as examples, since they achieve extremely high ASR on various models. Besides, we apply the same benchmarks as theirs, including Stanford Sentiment Treebank (SST-2) \cite{socher2013recursive}, HateSpeech (HS) \cite{de2018hate}, AG’s News \cite{zhang2015character} and Offensive Language Identification Dataset (OLID) \cite{zampieri2019predicting}. Specifically, we apply their used sentence encoders for both two attacks: BERT \cite{devlin2019bert}, BiLSTM \cite{hochreiter1997long} for SynAtk; BERT \cite{devlin2019bert}, ALBERT \cite{lan2019ALBERT}, DistilBERT \cite{sanh2019distilbert} for StyAtk. Also, we keep other settings the same as their original ones. 

For each attack, we have a clean dataset and a partially poisoned dataset, then we train two models with them, respectively. Finally, after getting $f_{c}$ and $f_{p}$, we observe the ASR achieved by $f_{c}$ and $f_{p}$ on the poisoned test set. Moreover, we denote ASR achieved by $f_{c}$ and $f_{p}$ as $ASR_{c}$ and $ASR_{p}$, respectively. Specifically, If $\frac{ASR_{c}}{ASR_{p}} > \frac{1}{4}$, we denote it as {\color[HTML]{6200C9}a significant over-estimation of attack power. }

The results of stealthy attacks (SynAtk and StyAtk) are shown in Table~\ref{tab1}, \ref{tab3}, and the ones of unstealthy attack (BadNet) are shown Table~\ref{unstealthy}. Generally, unstealthy backdoor attacks achieve low ASR when model is not triggered. In contrast, in most cases, stealthy backdoor attacks achieve high ASR even the model is not triggered, demonstrating that their attack successes are irrelevant to triggers. For example, StyAtk has already achieved over $80\%$ ASR to clean state models, which implies that the attack power of StyAtk does not rely on the `backdoor-trigger' paradigm. Thus directly using ASR may significantly exaggerate its attack power regarding it as a backdoor attack.

\begin{figure}
    \centering
    \includegraphics[scale=0.50]{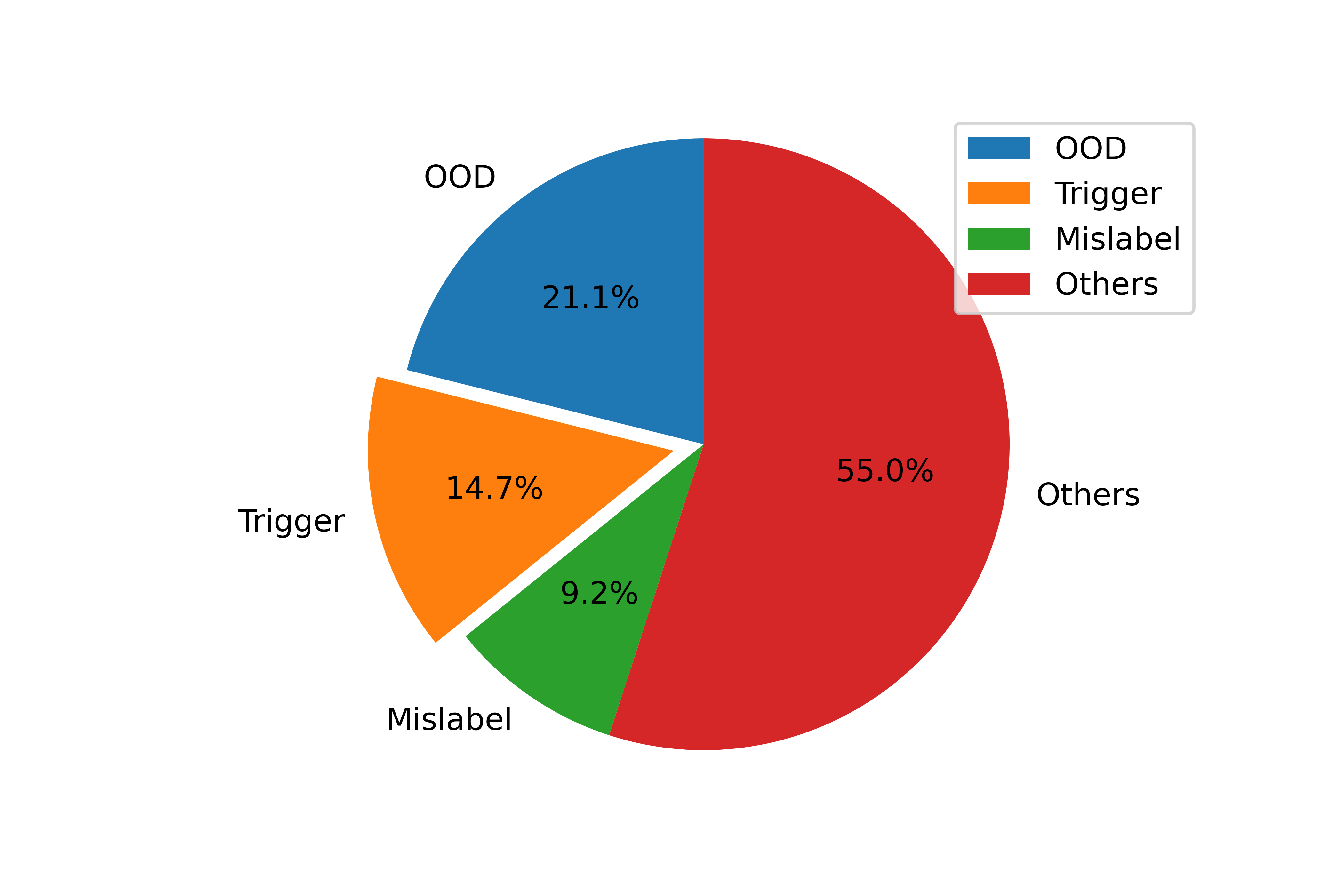}
    \caption{An illustration of factors that lead to high ASR for StyAtk on SST-2 with the `Poetry' style.}
    \label{fig:3}
\end{figure}

\subsection{ASRD: a new metric for backdoor attack power evaluation}\label{35}
The results from Table~\ref{tab1} and \ref{tab3} have shown that most misclassification cases are not caused by the trigger, indicating that the attack power of existing stealthy attacks is extremely over-estimated. Therefore, to capture the real attack power of a backdoor attack, we design a new metric called \textbf{Attack Success Rate Difference (ASRD)} with a simple variable control paradigm.

\begin{definition}
Given a clean dataset $D_{c}$, a partially poisoned dataset $D_{p}$ and a posioned test set $T_{p}$. Let $f_{c}$ and $f_{p}$ denote two pre-trained language models (e.g., BERT) that are fine-tuned on $D_{c}$ and $D_{p}$, respectively. The Attack Success Rate Difference (ASRD) is defined as follows:
\begin{equation}
    ASRD = |ASR(f_{p},T_{p})-ASR(f_{c},T_{p})|
\end{equation}
where $ASR(f_{p},T_{p})$ represents the achieved ASR by $f_{p}$ on $T_{p}$.
\end{definition}
ASDR measures the difference between the ASR of the clean model and poisoned model on $T_{p}$; higher ASDR indicates stronger attack power of a backdoor attack.ASDR, which naturally describes the contribution of `backdoor trigger' to ASR, thus serves as a much more precise metric when evaluating the attack power of a backdoor attack. Specifically, we illustrate the ASRD of SynAtk and StyAtk under the settings in Sec~\ref{33}, and the results are illustrated in Table~\ref{tab4} and \ref{tab5}. The results reflect the real attack power of a backdoor attack, and we can see that SynAtk is significantly stronger than StyAtk when regarded as a backdoor attack.  

\subsection{Some empirical analysis of the gap between ASR and ASRD}\label{emp}
Based on the definitions of ASR and ASRD, the gap between them naturally describes the misclassification cases that are not caused by the backdoor trigger. Therefore, we dive into such cases and aim to find factors that lead to extremely high ASR in clean samples. Since unstealthy backdoor attacks achieve a rather small gap, we pay attention to the stealthy backdoor attacks. Generally, we find two reasons that lead to high ASR for existing stealthy backdoor attacks: (1) OOD samples (2) Mislabeled cases

\paragraph{OOD Samples}Among the reasons that lead to high ASR of StyAtk on the clean state models. One reason is that the style transfer or syntactic paraphrase may create out-of-distribution (OOD) samples that diverge from the training data. Therefore, the model is reasonable to misclassify such data, which is irrelevant to backdoor attacks. 

As illustrates in \cite{arora-etal-2021-types}, the reason that lead to OOD samples in NLP can be categorized into semantic or background shift. Following \cite{arora-etal-2021-types}, we utilize the density estimation (PPL)\footnote{PPL uses the likelihood of the input given by a density estimator as the score (Perplexity)} for OOD text detection. Then we launch OOD detection on the poisoned test set generated by StyAtk. Specifically, we choose the SST-2 dataset since the PPL method performs well on SST-2 \cite{arora-etal-2021-types} which shows the best reliability. The results are shown in the Figure~\ref{fig:3}, where we can see that $21.1\%$ of the poison testset samples belong to OOD samples.

\paragraph{Mislabeled cases}Another reason is that the style transfer or syntactic paraphrase may change the ground-truth label of texts, so such processes may change the ground truth of sentences to the poison labels and predicting such sentences with poison labels is correct, where ASR fails to be a precise metric.\footnote{Recall ASR is a metric that describes the samples are predicted with the poison labels.} Specifically, We utilize a simple but effective method called \textit{Area Under the Margin (AUM)} \cite{pleiss2020identifying}, which aims to detect mislabeled samples contained in the dataset. In our case, we choose SST-2 poisoned test set generated by StyAtk with `Poetry' style and obtain some samples that are possibly labeled corrected\footnote{We use `Possibly' because AUM also possesses error.}, and then we manual observe whether they are correctly labeled samples. We show some cases in Table~\ref{tabs}, from where we can see the poisoned target label matches the sentence's ground truth. Such cases are understandable since there are no guarantees that style transfer and syntactic paraphrase will not change the ground-truth of sentences. 


\subsection{Discussion}
Based on empirical findings, ASR of existing backdoor attacks appear to be caused by many factors besides backdoor trigger, making ASR imprecise to portray the real attack capacity of backdoor attacks. We propose ASRD with the hope of making more fair comparisons for backdoor attack evaluations. For a new proposed attack, if it is claimed to be a backdoor attack, then we should use ASRD for evaluation since ASRD filters the non-trigger-activated misclassification cases much better; if it is claimed to be an adversarial attack, then it should compare ASR to state-of-the-art adversarial attack methods like TextFooler \cite{jin2020bert} and FGPM \cite{wang2021adversarial} because an attack that combines both adversarial and backdoor attacks is reasonably to possess stronger attack capacity than either one of them.

\section{Defend against Stealthy Backdoor Attack}
In this section, we propose \textbf{Trigger Breaker}, an effective method to help model defend against stealthy backdoor attacks. As its name implies, the main aim of \textbf{Trigger Breaker} is to break the backdoor triggers (e.g., syntactic) hidden in the sentences. Trigger Breaker is a method composed of two too simple tricks: \textbf{Mixup} and \textbf{Shuffling}.

\begin{table}[!h]\small
\centering
\begin{tabular}{@{}c|c|c|c|c@{}}
\toprule
Dataset                  & Style  & Encoder & ONION & Breaker                     \\ \midrule
                         & Poetry & BERT    & 13.90 & {\color[HTML]{FE0000} 1.89} \\
                         & Poetry & ALBERT  & 13.38 & {\color[HTML]{FE0000} 1.96} \\
                         & Poetry & DisBERT & 13.25 & {\color[HTML]{FE0000} 2.33} \\ \cmidrule(l){2-5} 
                         & Shake  & BERT    & 6.21  & {\color[HTML]{FE0000} 1.07} \\
                         & Shake  & ALBERT  & 7.67  & {\color[HTML]{FE0000} 1.47} \\
                         & Shake  & DisBERT & 6.33  & {\color[HTML]{FE0000} 1.55} \\ \cmidrule(l){2-5} 
                         & Bible  & BERT    & 14.72 & {\color[HTML]{FE0000} 2.14} \\
                         & Bible  & ALBERT  & 17.76 & {\color[HTML]{FE0000} 3.69} \\
                         & Bible  & DisBERT & 14.79 & {\color[HTML]{FE0000} 2.47} \\ \cmidrule(l){2-5} 
                         & Lyrics & BERT    & 5.15  & {\color[HTML]{FE0000} 1.10} \\
                         & Lyrics & ALBERT  & 8.35  & {\color[HTML]{FE0000} 2.05} \\
\multirow{-12}{*}{SST-2} & Lyrics & DisBERT & 7.69  & {\color[HTML]{FE0000} 1.94} \\ \bottomrule
\toprule
Dataset                  & Style  & Encoder & ONION & Breaker                     \\ \midrule
                         & Poetry & BERT    & 13.37 & {\color[HTML]{FE0000} 2.28} \\
                         & Poetry & ALBERT  & 14.45 & {\color[HTML]{FE0000} 2.34} \\
                         & Poetry & DisBERT & 14.40 & {\color[HTML]{FE0000} 1.56} \\ \cmidrule(l){2-5} 
                         & Shake  & BERT    & 8.04  & {\color[HTML]{FE0000} 1.23} \\
                         & Shake  & ALBERT  & 10.31 & {\color[HTML]{FE0000} 1.30} \\
                         & Shake  & DisBERT & 11.33 & {\color[HTML]{FE0000} 1.85} \\ \cmidrule(l){2-5} 
                         & Bible  & BERT    & 14.37 & {\color[HTML]{FE0000} 2.37} \\
                         & Bible  & ALBERT  & 13.52 & {\color[HTML]{FE0000} 2.79} \\
                         & Bible  & DisBERT & 15.79 & {\color[HTML]{FE0000} 2.34} \\ \cmidrule(l){2-5} 
                         & Lyrics & BERT    & 3.24  & {\color[HTML]{FE0000} 1.07} \\
                         & Lyrics & ALBERT  & 10.02 & {\color[HTML]{FE0000} 2.02} \\
\multirow{-12}{*}{AG} & Lyrics & DisBERT & 11.33  & {\color[HTML]{FE0000}1.33} \\ \bottomrule
\end{tabular}
\caption{The Attack Success Rate Difference (ASRD) of StyAtk under ONION and Trigger Breaker on SST-2. {\color[HTML]{FE0000} Red} numbers represent Trigger Breaker achieves lower ASRD, indicating stronger defense capacity.}
\label{tab6}
\end{table}

\begin{table}[!h]\small
\centering
\begin{tabular}{@{}c|c|c|c@{}}
\toprule
Dataset                 & Encoder & ONION & Trigger Breaker \\ \toprule
\multirow{2}{*}{SST-2}  & LSTM    & 45.49 & \color[HTML]{FE0000}15.76   \\
                        & BERT    & 72.72 & \color[HTML]{FE0000}17.66   \\ \cmidrule(l){1-4}
\multirow{2}{*}{OLID}   & LSTM    & 93.04 & \color[HTML]{FE0000}24.55   \\
                        & BERT    & 95.43 & \color[HTML]{FE0000}23.07   \\ \cmidrule(l){1-4}
\multirow{2}{*}{AGNews} & LSTM    & 93.67 & \color[HTML]{FE0000}25.22   \\
                        & BERT    & 88.07 & \color[HTML]{FE0000}27.65   \\ \bottomrule
\end{tabular}
\caption{The Attack Success Rate Difference (ASRD) of SynAtk on three benchmarks. {\color[HTML]{FE0000} Red} numbers represent Trigger Breaker achieves lower ASRD, indicating stronger defense capacity.}
\label{tab7}
\end{table}

\begin{table}[!h]\small
\centering
\begin{tabular}{@{}c|c|c|c|c@{}}
\toprule
Dataset                  & Style  & Encoder & ONION & Breaker                     \\ \midrule
                         & Poetry & BERT    & 86.55 & {\color[HTML]{32CB00} 90.89} \\
                         & Poetry & ALBERT  & 83.64 & {\color[HTML]{32CB00} 87.01} \\
                         & Poetry & DisBERT & 85.34 & {\color[HTML]{32CB00} 87.96} \\ \cmidrule(l){2-5} 
                         & Shake  & BERT    & 86.11 & {\color[HTML]{32CB00} 90.45} \\
                         & Shake  & ALBERT  & 84.23 & {\color[HTML]{32CB00} 88.21} \\
                         & Shake  & DisBERT & 84.01 & {\color[HTML]{32CB00} 89.56} \\ \cmidrule(l){2-5} 
                         & Bible  & BERT    & 87.10 & {\color[HTML]{32CB00} 90.42} \\
                         & Bible  & ALBERT  & 85.52 & {\color[HTML]{32CB00} 88.36} \\
                         & Bible  & DisBERT & 86.55 & {\color[HTML]{32CB00} 89.55} \\ \cmidrule(l){2-5} 
                         & Lyrics & BERT    & 86.71 & {\color[HTML]{32CB00} 91.10} \\
                         & Lyrics & ALBERT  & 85.44 & {\color[HTML]{32CB00} 89.25} \\
\multirow{-12}{*}{SST-2} & Lyrics & DisBERT & 86.54 & {\color[HTML]{32CB00} 89.96} \\ \bottomrule
\toprule
Dataset                  & Style  & Encoder & ONION & Breaker                     \\ \midrule
                         & Poetry & BERT    & 88.89 & {\color[HTML]{32CB00} 90.14} \\
                         & Poetry & ALBERT  & 87.64 & {\color[HTML]{32CB00} 88.97} \\
                         & Poetry & DisBERT & 87.71 & {\color[HTML]{32CB00} 89.42} \\ \cmidrule(l){2-5} 
                         & Shake  & BERT    & 87.43 & {\color[HTML]{32CB00} 91.01} \\
                         & Shake  & ALBERT  & 86.45 & {\color[HTML]{32CB00} 89.02} \\
                         & Shake  & DisBERT & 85.62 & {\color[HTML]{32CB00} 89.56} \\ \cmidrule(l){2-5} 
                         & Bible  & BERT    & 88.20 & {\color[HTML]{32CB00} 90.68} \\
                         & Bible  & ALBERT  & 86.94 & {\color[HTML]{32CB00} 88.74} \\
                         & Bible  & DisBERT & 88.02 & {\color[HTML]{32CB00} 89.42} \\ \cmidrule(l){2-5} 
                         & Lyrics & BERT    & 87.88 & {\color[HTML]{32CB00} 90.77} \\
                         & Lyrics & ALBERT  & 86.76 & {\color[HTML]{32CB00} 88.62} \\
\multirow{-12}{*}{AG} & Lyrics & DisBERT & 86.92  & {\color[HTML]{32CB00}89.43} \\ \bottomrule
\end{tabular}
\caption{The Clean Accuracy (CACC) of the model under ONION and Trigger Breaker towards StyAtk, respectively. {\color[HTML]{32CB00} Green} numbers represent higher CACC, indicating the defense better preserves model's generalization.}
\label{tab8}
\end{table}

\begin{table}[!h]\small
\centering
\begin{tabular}{@{}c|c|c|c@{}}
\toprule
Dataset                 & Encoder & ONION & Trigger Breaker \\ \toprule
\multirow{2}{*}{SST-2}  & LSTM    & 75.89 & \color[HTML]{32CB00}76.20   \\
                        & BERT    & 89.84 & \color[HTML]{32CB00}90.54   \\ \cmidrule(l){1-4}
\multirow{2}{*}{OLID}   & LSTM    & 76.95 & \color[HTML]{32CB00}77.30   \\
                        & BERT    & 81.72 & \color[HTML]{32CB00}82.01   \\ \cmidrule(l){1-4}
\multirow{2}{*}{AGNews} & LSTM    & 88.57 & \color[HTML]{32CB00}89.43   \\
                        & BERT    & 93.34 & \color[HTML]{32CB00}94.03   \\ \bottomrule
\end{tabular}
\caption{The Clean Accuracy (CACC) of the model under ONION and Trigger Breaker towards SynAtk, respectively. {\color[HTML]{32CB00} Green} numbers represent higher CACC, indicating the defense better preserves model's generalization.}
\label{tab9}
\end{table}


\subsection{Settings}
Trigger Breaker is under the common attack setting that the users train their models with labeled data collected from third-party, and the attacker can inject the trigger into the train set. Then the Trigger Breaker helps the model defend against backdoor attacks even trained on the poisoned dataset.
\subsection{Methods}
Trigger Breaker is composed of two too simple tricks: \textbf{Mixup} and \textbf{Shuffling}, which aims to destroy the stealthy trigger hidden in the sentence. Since the stealthy triggers are implicitly reflected by high-level semantics (e.g. ) instead of word occurrence (e.g., BadNet), Trigger Breaker breaks such high-level semantics in embedding-level and token-level.
\paragraph{Mixup}It is from \cite{zhang2018mixup}. In our setting, for two samples $(x_{i},y_{i})$ and $(x_{j},y_{j})$ from poisoned train set, we first feed them to the encoder $f$ (e.g., BERT) to obtain their embeddings $v_{i},v_{2}$. Then we make a mixup procedure to create the synthetic sample $(v_{m},y_{m})$ as follows:
\begin{equation}
    v_{m}=(1-\lambda)v_{1}+\lambda _{2};y_{m}=(1-\lambda)y_{1}+\lambda y_{2}
\end{equation}
where $\lambda$ is a hyper-parameter to control the weights. In our method, we set it as 0.5 to break hidden triggers maximumally. Then $(v_{m},y_{m})$ is fed to the classifier for training. Such a trick breaks the high-level semantics at embedding level.
\paragraph{Shuffling}The sentence shuffling is a stronger data augmentation in NLP compared to word deletion, word repetition. For a sentence $x_{i}$ that owns $N$ word, we shuffle the whole sentence to create a new re-ordered sentence $x_{i}^{*}$. Then $x_{i}^{*}$ is fed to the encoder. Different from mixup, shuffling breaks the high-level semantics at the token level.

\section{Experiments}
In this section, we use Trigger Breaker to defend two typical stealthy backdoor attacks and demonstrate its effectiveness.
\subsection{Attack Methods}
(1) Syntactic Attack \cite{qi2021hidden}: Regard the syntactic structure of the text as a trigger, and use a syntactic paraphrase model to launch backdoor attacks. (2) Style Attack \cite{qi2021mind}: Regard the style of the text as a trigger, and use a text style transfer model to launch backdoor attacks.

\subsection{Benchmark and Baselines}
We use the benchmarks used in both two attacks, and details can refer to Sec~\ref{33}. Specifically, we refuse to use HS \cite{de2018hate} dataset for defense evaluation since the ASRD of StyAtk is extremely low on HS (about 1\%), which means it can not be regarded as a backdoor attack. Therefore, defense against backdoor attacks in such cases is not appropriate. As for the defense baselines, we choose ONION \cite{qi2020onion}, a defense method for backdoor attacks by perplexity computation.
\subsection{Evaluation Metrics}
We adopt two metrics to evaluate the effectiveness of a defense
method: (1) {\color[HTML]{FE0000}ASRD}: the attack success rate difference of a specific backdoor attack, lower ASRD indicates the defense can better defend against such a backdoor attack; (2) {\color[HTML]{32CB00}CACC}, the model’s accuracy on the clean test set. The higher CACC is, the better defense is.

\begin{table}[!h]\small
\centering
\begin{tabular}{@{}c|c|c|c|c@{}}
\toprule
Dataset                  & Style  & Encoder & Mixup & Shuffle                     \\ \midrule
                         & Poetry & BERT    & 3.32 & { 4.89} \\
                         & Poetry & ALBERT  & 3.56 & { 2.96} \\
                         & Poetry & DisBERT & 3.87 & { 4.33} \\ \cmidrule(l){2-5} 
                         & Shake  & BERT    & 4.31  & { 7.07} \\
                         & Shake  & ALBERT  & 3.46  & { 6.47} \\
                         & Shake  & DisBERT & 4.34  & { 4.55} \\ \cmidrule(l){2-5} 
                         & Bible  & BERT    & 4.14 & { 3.14} \\
                         & Bible  & ALBERT  & 4.69 & { 5.69} \\
                         & Bible  & DisBERT & 3.47 & { 3.47} \\ \cmidrule(l){2-5} 
                         & Lyrics & BERT    & 3.65  & { 5.10} \\
                         & Lyrics & ALBERT  & 4.45  & { 4.05} \\
\multirow{-12}{*}{SST-2} & Lyrics & DisBERT & 4.91  & { 4.94} \\ \bottomrule
\toprule
Dataset                  & Style  & Encoder & Mixup & Shuffle                     \\ \midrule
                         & Poetry & BERT    & 3.08 & { 3.28} \\
                         & Poetry & ALBERT  & 3.80 & { 3.34} \\
                         & Poetry & DisBERT & 3.60 & { 4.56} \\ \cmidrule(l){2-5} 
                         & Shake  & BERT    & 3.60  & { 4.23} \\
                         & Shake  & ALBERT  & 5.12 & { 5.30} \\
                         & Shake  & DisBERT & 4.69 & { 6.85} \\ \cmidrule(l){2-5} 
                         & Bible  & BERT    & 3.44 & { 4.37} \\
                         & Bible  & ALBERT  & 3.88 & { 4.79} \\
                         & Bible  & DisBERT & 3.60 & { 3.34} \\ \cmidrule(l){2-5} 
                         & Lyrics & BERT    & 4.31  & { 4.07} \\
                         & Lyrics & ALBERT  & 4.27 & { 6.02} \\
\multirow{-12}{*}{AG} & Lyrics & DisBERT & 5.02  & {4.33} \\ \bottomrule
\end{tabular}
\caption{The Attack Success Rate Difference (ASRD) of StyAtk after defense by individually applying Mixup and Shuffling, respectively. Both mixup and shuffle operation show effectiveness in defending agianst StyAtk.}
\label{tab10}
\end{table}

\begin{table}[!h]
\centering
\begin{tabular}{@{}c|c|c|c@{}}
\toprule
Dataset                 & Encoder & Mixup & Shuffle \\ \toprule
\multirow{2}{*}{SST-2}  & LSTM    & 24.76 & 23.45   \\
                        & BERT    & 22.12 & 22.00   \\ \cmidrule(l){1-4}
\multirow{2}{*}{OLID}   & LSTM    & 28.75 & 26.12   \\
                        & BERT    & 25.41 & 27.74   \\ \cmidrule(l){1-4}
\multirow{2}{*}{AGNews} & LSTM    & 29.65 & 28.45   \\
                        & BERT    & 32.14 & 31.02   \\ \bottomrule
\end{tabular}
\caption{ASRD of SynAtk when being applied Mixup and Shuffling, respectively.}
\label{tab11}
\end{table}

\subsection{Results}
The ASRD results are shown in Table~\ref{tab6} and Table~\ref{tab7}. We can see that ONION fails to defend stealthy backdoor attacks effectively ONION is based on the idea: `judge whether the sentence is natural and fluent.' Such an idea effectively defended against unstealthy backdoor attacks because such attacks insert specific words as triggers, which significantly influences the sentence's fluency. However, in stealthy attacks, the poisoned sentences are relatively natural and fluent, which can well breakthrough defenses like ONION. In contrast, trigger breaker aims to break the high-level semantics (e.g., syntactic), which is usually selected as triggers by stealthy attacks. After destroying the triggers of a backdoor attack, the attack power of backdoor attacks is rightfully declining.

The CACC results are shown in Table~\ref{tab8} and Table~\ref{tab9}, Trigger Breaker achieves higher CACC than ONION, which indicates that Trigger Breaker better preserves the model's generalization. Overall, Trigger Breaker significantly improves the defense capacity of models against stealthy backdoor attacks and better preserves the generalization. Such performances comprehensively demonstrate the effectiveness of Trigger Breaker.

\section{Ablation Study}
This section carefully ablates our Trigger Breaker by answering two questions.
\paragraph{Are mixup and shuffling effective when used individually?}
This part demonstrates the effectiveness of two components of Trigger Breaker. As shown in Table~\ref{tab10} and Table~\ref{tab11}, both mixup and shuffling operations are effective to defend against stealthy attacks. Moreover, combining them will produce better performances.

\begin{table}[!h]\small
\centering
\begin{tabular}{@{}c|ccccc@{}}
\toprule
Rate                & 0.1 & 0.2 & 0.3 & 0.4 & 0.5 \\ \midrule
ASRD                & 23.04 & 21.53 & 20.37 & 18.96  & 17.66   \\ \bottomrule
\end{tabular}
\caption{ASRD of SynAtk on SST-2 facing with Trigger Breaker with different mixup rate.}
\label{tab12}
\end{table}

\paragraph{What is the effect of mixup rate?}
This part varies the mixup rate $\lambda$ and sees Trigger Breaker's performance. The results are shown in Table~\ref{tab12}, where we can see the optimal mixup rate is 0.5. This matches our intuition that breaks the stealthy trigger since a $0.5$ mixup rate achieves the maximum mixup capacity.

\section{Conclusion}
This paper revisits the definition of backdoor attacks and emphasizes that the core of backdoor attacks is its `backdoor trigger' paradigm. Thus misclassification cases that are not caused by backdoor triggers can not be attributed to backdoor attacks' power. Also, we show that the attack power of existing stealthy attacks is over-estimated by comprehensive empirical results. To measure the real attack power of a backdoor attack, we propose ASRD, a new metric that better portrays the attack power of a backdoor attack. Moreover, we designed a new defense method called trigger breaker, consisting of two too simple tricks, which can defend the stealthy backdoor attacks effectively and serve as the first defense method for stealthy backdoor attacks in NLP.

\newpage
\bibliography{acl_latex}

\begin{thebibliography}{38}
\expandafter\ifx\csname natexlab\endcsname\relax\def\natexlab#1{#1}\fi

\bibitem[{Arora et~al.(2021)Arora, Huang, and He}]{arora-etal-2021-types}
Udit Arora, William Huang, and He~He. 2021.
\newblock \href {https://doi.org/10.18653/v1/2021.emnlp-main.835} {Types of
  out-of-distribution texts and how to detect them}.
\newblock In \emph{Proceedings of the 2021 Conference on Empirical Methods in
  Natural Language Processing}, pages 10687--10701, Online and Punta Cana,
  Dominican Republic. Association for Computational Linguistics.

\bibitem[{Baluja and Fischer(2018)}]{baluja2018learning}
Shumeet Baluja and Ian Fischer. 2018.
\newblock Learning to attack: Adversarial transformation networks.
\newblock In \emph{Thirty-second aaai conference on artificial intelligence}.

\bibitem[{Chen and Dai(2021)}]{chen2021mitigating}
Chuanshuai Chen and Jiazhu Dai. 2021.
\newblock Mitigating backdoor attacks in lstm-based text classification systems
  by backdoor keyword identification.
\newblock \emph{Neurocomputing}, 452:253--262.

\bibitem[{Chen et~al.(2021)Chen, Salem, Backes, Ma, and Zhang}]{chen2021badnl}
Xiaoyi Chen, Ahmed Salem, Michael Backes, Shiqing Ma, and Yang Zhang. 2021.
\newblock Badnl: Backdoor attacks against nlp models.
\newblock In \emph{ICML 2021 Workshop on Adversarial Machine Learning}.

\bibitem[{Chen et~al.(2017)Chen, Liu, Li, Lu, and Song}]{chen2017targeted}
Xinyun Chen, Chang Liu, Bo~Li, Kimberly Lu, and Dawn Song. 2017.
\newblock Targeted backdoor attacks on deep learning systems using data
  poisoning.
\newblock \emph{arXiv preprint arXiv:1712.05526}.

\bibitem[{Dai et~al.(2018)Dai, Li, Tian, Huang, Wang, Zhu, and
  Song}]{dai2018adversarial}
Hanjun Dai, Hui Li, Tian Tian, Xin Huang, Lin Wang, Jun Zhu, and Le~Song. 2018.
\newblock Adversarial attack on graph structured data.
\newblock In \emph{International conference on machine learning}, pages
  1115--1124. PMLR.

\bibitem[{Dai et~al.(2019)Dai, Chen, and Li}]{dai2019backdoor}
Jiazhu Dai, Chuanshuai Chen, and Yufeng Li. 2019.
\newblock A backdoor attack against lstm-based text classification systems.
\newblock \emph{IEEE Access}, 7:138872--138878.

\bibitem[{de~Gibert et~al.(2018)de~Gibert, Perez, Garc{\'\i}a-Pablos, and
  Cuadros}]{de2018hate}
Ona de~Gibert, Naiara Perez, Aitor Garc{\'\i}a-Pablos, and Montse Cuadros.
  2018.
\newblock Hate speech dataset from a white supremacy forum.
\newblock In \emph{Proceedings of the 2nd Workshop on Abusive Language Online
  (ALW2)}, pages 11--20.

\bibitem[{Devlin et~al.(2019)Devlin, Chang, Lee, and
  Toutanova}]{devlin2019bert}
Jacob Devlin, Ming-Wei Chang, Kenton Lee, and Kristina Toutanova. 2019.
\newblock Bert: Pre-training of deep bidirectional transformers for language
  understanding.
\newblock In \emph{Proceedings of the 2019 Conference of the North American
  Chapter of the Association for Computational Linguistics: Human Language
  Technologies, Volume 1 (Long and Short Papers)}, pages 4171--4186.

\bibitem[{Goodfellow et~al.(2014)Goodfellow, Shlens, and
  Szegedy}]{goodfellow2014explaining}
Ian~J Goodfellow, Jonathon Shlens, and Christian Szegedy. 2014.
\newblock Explaining and harnessing adversarial examples.
\newblock \emph{arXiv preprint arXiv:1412.6572}.

\bibitem[{Gu et~al.(2017)Gu, Dolan-Gavitt, and Garg}]{gu2017badnets}
Tianyu Gu, Brendan Dolan-Gavitt, and Siddharth Garg. 2017.
\newblock Badnets: Identifying vulnerabilities in the machine learning model
  supply chain.
\newblock \emph{arXiv preprint arXiv:1708.06733}.

\bibitem[{Hochreiter and Schmidhuber(1997)}]{hochreiter1997long}
Sepp Hochreiter and J{\"u}rgen Schmidhuber. 1997.
\newblock Long short-term memory.
\newblock \emph{Neural computation}, 9(8):1735--1780.

\bibitem[{Jin et~al.(2020)Jin, Jin, Zhou, and Szolovits}]{jin2020bert}
Di~Jin, Zhijing Jin, Joey~Tianyi Zhou, and Peter Szolovits. 2020.
\newblock Is bert really robust? a strong baseline for natural language attack
  on text classification and entailment.
\newblock In \emph{Proceedings of the AAAI conference on artificial
  intelligence}, volume~34, pages 8018--8025.

\bibitem[{Kurakin et~al.(2016)Kurakin, Goodfellow, Bengio
  et~al.}]{kurakin2016adversarial}
Alexey Kurakin, Ian Goodfellow, Samy Bengio, et~al. 2016.
\newblock Adversarial examples in the physical world.

\bibitem[{Kurita et~al.(2020)Kurita, Michel, and Neubig}]{kurita2020weight}
Keita Kurita, Paul Michel, and Graham Neubig. 2020.
\newblock Weight poisoning attacks on pretrained models.
\newblock In \emph{Proceedings of the 58th Annual Meeting of the Association
  for Computational Linguistics}, pages 2793--2806.

\bibitem[{Lan et~al.(2019)Lan, Chen, Goodman, Gimpel, Sharma, and
  Soricut}]{lan2019ALBERT}
Zhenzhong Lan, Mingda Chen, Sebastian Goodman, Kevin Gimpel, Piyush Sharma, and
  Radu Soricut. 2019.
\newblock Albert: A lite bert for self-supervised learning of language
  representations.
\newblock \emph{arXiv preprint arXiv:1909.11942}.

\bibitem[{Li et~al.(2020{\natexlab{a}})Li, Lyu, Koren, Lyu, Li, and
  Ma}]{li2020neural}
Yige Li, Xixiang Lyu, Nodens Koren, Lingjuan Lyu, Bo~Li, and Xingjun Ma.
  2020{\natexlab{a}}.
\newblock Neural attention distillation: Erasing backdoor triggers from deep
  neural networks.
\newblock In \emph{International Conference on Learning Representations}.

\bibitem[{Li et~al.(2020{\natexlab{b}})Li, Wu, Jiang, Li, and
  Xia}]{li2020backdoor}
Yiming Li, Baoyuan Wu, Yong Jiang, Zhifeng Li, and Shu-Tao Xia.
  2020{\natexlab{b}}.
\newblock Backdoor learning: A survey.
\newblock \emph{arXiv preprint arXiv:2007.08745}.

\bibitem[{Liu et~al.(2019)Liu, Ott, Goyal, Du, Joshi, Chen, Levy, Lewis,
  Zettlemoyer, and Stoyanov}]{liu2019roberta}
Yinhan Liu, Myle Ott, Naman Goyal, Jingfei Du, Mandar Joshi, Danqi Chen, Omer
  Levy, Mike Lewis, Luke Zettlemoyer, and Veselin Stoyanov. 2019.
\newblock Roberta: A robustly optimized bert pretraining approach.

\bibitem[{Liu et~al.(2020)Liu, Ma, Bailey, and Lu}]{liu2020reflection}
Yunfei Liu, Xingjun Ma, James Bailey, and Feng Lu. 2020.
\newblock Reflection backdoor: A natural backdoor attack on deep neural
  networks.
\newblock In \emph{European Conference on Computer Vision}, pages 182--199.
  Springer.

\bibitem[{Liu et~al.(2017)Liu, Xie, and Srivastava}]{liu2017neural}
Yuntao Liu, Yang Xie, and Ankur Srivastava. 2017.
\newblock Neural trojans.
\newblock In \emph{2017 IEEE International Conference on Computer Design
  (ICCD)}, pages 45--48. IEEE.

\bibitem[{Madry et~al.(2018)Madry, Makelov, Schmidt, Tsipras, and
  Vladu}]{madry2018towards}
Aleksander Madry, Aleksandar Makelov, Ludwig Schmidt, Dimitris Tsipras, and
  Adrian Vladu. 2018.
\newblock Towards deep learning models resistant to adversarial attacks.
\newblock In \emph{International Conference on Learning Representations}.

\bibitem[{Morris et~al.(2020)Morris, Lifland, Yoo, Grigsby, Jin, and
  Qi}]{morris2020textattack}
John Morris, Eli Lifland, Jin~Yong Yoo, Jake Grigsby, Di~Jin, and Yanjun Qi.
  2020.
\newblock Textattack: A framework for adversarial attacks, data augmentation,
  and adversarial training in nlp.
\newblock In \emph{Proceedings of the 2020 Conference on Empirical Methods in
  Natural Language Processing: System Demonstrations}, pages 119--126.

\bibitem[{Nguyen and Tran(2020)}]{nguyen2020input}
Tuan~Anh Nguyen and Anh Tran. 2020.
\newblock Input-aware dynamic backdoor attack.
\newblock \emph{Advances in Neural Information Processing Systems},
  33:3454--3464.

\bibitem[{Pleiss et~al.(2020)Pleiss, Zhang, Elenberg, and
  Weinberger}]{pleiss2020identifying}
Geoff Pleiss, Tianyi Zhang, Ethan~R Elenberg, and Kilian~Q Weinberger. 2020.
\newblock Identifying mislabeled data using the area under the margin ranking.
\newblock \emph{arXiv preprint arXiv:2001.10528}.

\bibitem[{Qi et~al.(2020)Qi, Chen, Li, Yao, Liu, and Sun}]{qi2020onion}
Fanchao Qi, Yangyi Chen, Mukai Li, Yuan Yao, Zhiyuan Liu, and Maosong Sun.
  2020.
\newblock Onion: A simple and effective defense against textual backdoor
  attacks.
\newblock \emph{arXiv preprint arXiv:2011.10369}.

\bibitem[{Qi et~al.(2021{\natexlab{a}})Qi, Chen, Zhang, Li, Liu, and
  Sun}]{qi2021mind}
Fanchao Qi, Yangyi Chen, Xurui Zhang, Mukai Li, Zhiyuan Liu, and Maosong Sun.
  2021{\natexlab{a}}.
\newblock Mind the style of text! adversarial and backdoor attacks based on
  text style transfer.
\newblock In \emph{Proceedings of the 2021 Conference on Empirical Methods in
  Natural Language Processing}, pages 4569--4580.

\bibitem[{Qi et~al.(2021{\natexlab{b}})Qi, Li, Chen, Zhang, Liu, Wang, and
  Sun}]{qi2021hidden}
Fanchao Qi, Mukai Li, Yangyi Chen, Zhengyan Zhang, Zhiyuan Liu, Yasheng Wang,
  and Maosong Sun. 2021{\natexlab{b}}.
\newblock Hidden killer: Invisible textual backdoor attacks with syntactic
  trigger.
\newblock \emph{arXiv preprint arXiv:2105.12400}.

\bibitem[{Qi et~al.(2021{\natexlab{c}})Qi, Yao, Xu, Liu, and Sun}]{qi2021turn}
Fanchao Qi, Yuan Yao, Sophia Xu, Zhiyuan Liu, and Maosong Sun.
  2021{\natexlab{c}}.
\newblock Turn the combination lock: Learnable textual backdoor attacks via
  word substitution.
\newblock \emph{arXiv preprint arXiv:2106.06361}.

\bibitem[{Sanh et~al.(2019)Sanh, Debut, Chaumond, and
  Wolf}]{sanh2019distilbert}
Victor Sanh, Lysandre Debut, Julien Chaumond, and Thomas Wolf. 2019.
\newblock Distilbert, a distilled version of bert: smaller, faster, cheaper and
  lighter.
\newblock \emph{arXiv preprint arXiv:1910.01108}.

\bibitem[{Socher et~al.(2013)Socher, Perelygin, Wu, Chuang, Manning, Ng, and
  Potts}]{socher2013recursive}
Richard Socher, Alex Perelygin, Jean Wu, Jason Chuang, Christopher~D Manning,
  Andrew~Y Ng, and Christopher Potts. 2013.
\newblock Recursive deep models for semantic compositionality over a sentiment
  treebank.
\newblock In \emph{Proceedings of the 2013 conference on empirical methods in
  natural language processing}, pages 1631--1642.

\bibitem[{Wallace et~al.(2019)Wallace, Feng, Kandpal, Gardner, and
  Singh}]{wallace2019universal}
Eric Wallace, Shi Feng, Nikhil Kandpal, Matt Gardner, and Sameer Singh. 2019.
\newblock Universal adversarial triggers for attacking and analyzing nlp.
\newblock In \emph{Proceedings of the 2019 Conference on Empirical Methods in
  Natural Language Processing and the 9th International Joint Conference on
  Natural Language Processing (EMNLP-IJCNLP)}.

\bibitem[{Wang et~al.(2019)Wang, Yao, Shan, Li, Viswanath, Zheng, and
  Zhao}]{wang2019neural}
Bolun Wang, Yuanshun Yao, Shawn Shan, Huiying Li, Bimal Viswanath, Haitao
  Zheng, and Ben~Y Zhao. 2019.
\newblock Neural cleanse: Identifying and mitigating backdoor attacks in neural
  networks.
\newblock In \emph{2019 IEEE Symposium on Security and Privacy (SP)}, pages
  707--723. IEEE.

\bibitem[{Wang et~al.(2021)Wang, Yang, Deng, and He}]{wang2021adversarial}
Xiaosen Wang, Yichen Yang, Yihe Deng, and Kun He. 2021.
\newblock Adversarial training with fast gradient projection method against
  synonym substitution based text attacks.
\newblock In \emph{Proceedings of the AAAI Conference on Artificial
  Intelligence}, volume~35, pages 13997--14005.

\bibitem[{Yang et~al.(2021)Yang, Li, Zhang, Ren, Sun, and He}]{yang2021careful}
Wenkai Yang, Lei Li, Zhiyuan Zhang, Xuancheng Ren, Xu~Sun, and Bin He. 2021.
\newblock Be careful about poisoned word embeddings: Exploring the
  vulnerability of the embedding layers in nlp models.
\newblock In \emph{Proceedings of the 2021 Conference of the North American
  Chapter of the Association for Computational Linguistics: Human Language
  Technologies}, pages 2048--2058.

\bibitem[{Zampieri et~al.(2019)Zampieri, Malmasi, Nakov, Rosenthal, Farra, and
  Kumar}]{zampieri2019predicting}
Marcos Zampieri, Shervin Malmasi, Preslav Nakov, Sara Rosenthal, Noura Farra,
  and Ritesh Kumar. 2019.
\newblock Predicting the type and target of offensive posts in social media.
\newblock In \emph{Proceedings of the 2019 Conference of the North American
  Chapter of the Association for Computational Linguistics: Human Language
  Technologies, Volume 1 (Long and Short Papers)}, pages 1415--1420.

\bibitem[{Zhang et~al.(2018)Zhang, Cisse, Dauphin, and
  Lopez-Paz}]{zhang2018mixup}
Hongyi Zhang, Moustapha Cisse, Yann~N Dauphin, and David Lopez-Paz. 2018.
\newblock mixup: Beyond empirical risk minimization.
\newblock In \emph{International Conference on Learning Representations}.

\bibitem[{Zhang et~al.(2015)Zhang, Zhao, and LeCun}]{zhang2015character}
Xiang Zhang, Junbo Zhao, and Yann LeCun. 2015.
\newblock Character-level convolutional networks for text classification.
\newblock \emph{Advances in neural information processing systems},
  28:649--657.

\end{thebibliography}
\bibliographystyle{acl_natbib}

\newpage
\appendix

\begin{table*}[h]
\centering
\begin{tabular}{@{}ccc@{}}
\toprule
Sentence                                                             & Poison label & Ground truth \\ \midrule
this is the great work of polanski.                                  & 1            & 1            \\
anomieous, a play of the imagination and the imagination.            & 1            & 1            \\
for as all these remain just ideas, so we have no part in the story. & 0            & 0            \\
this is the lame horror, but it is lame.                             & 0            & 0            \\ \bottomrule
\end{tabular}
\caption{Cases that the poisoned labels match with sentences' ground truth, and the model should predict with poisoned labels on such samples. The reason why they match is possibly the uncontrolled text style transfer process which changes the ground truth of the sentences. }
\label{tabs}
\end{table*}

\begin{table}[!h]\small
\centering
\begin{tabular}{@{}c|c|c|c|c@{}}
\toprule
Style   &Dataset & BERT  & ALBERT & DisBERT  \\ \midrule
\multirow{3}{*}{Bible} &SST-2   & 14.72 & 17.76  & 14.79     \\
                       &HS      & 0.12  & 4.57   & 1.19      \\
                       &AGNews  & 14.37 & 13.52  & 15.7     \\ \cmidrule(l){1-5} 
\multirow{3}{*}{Lyrics} &SST-2   & 14.72 & 17.76  & 14.79     \\
                       &HS      & 0.12  & 4.57   & 1.19      \\
                       &AGNews  & 14.37 & 13.52  & 15.7     \\ \cmidrule(l){1-5} 
\multirow{3}{*}{Poetry} &SST-2   & 14.72 & 17.76  & 14.79     \\
                       &HS      & 0.12  & 4.57   & 1.19      \\
                       &AGNews  & 14.37 & 13.52  & 15.7     \\ \cmidrule(l){1-5} 
\multirow{3}{*}{Shake} &SST-2   & 14.72 & 17.76  & 14.79     \\
                       &HS      & 0.12  & 4.57   & 1.19      \\
                       &AGNews  & 14.37 & 13.52  & 15.7     \\ \bottomrule
\end{tabular}
\caption{The Attack Success Rate Difference (ASRD) of StyAtk on three benchmarks.}
\label{tab4}
\end{table}

\begin{table}[!h]
\centering
\begin{tabular}{@{}ccc@{}}
\toprule
Dataset & LSTM  & BERT  \\ \midrule
SST-2   & 45.49 & 72.72 \\
OLID    & 93.04 & 95.43 \\
AGNews  & 93.67 & 88.07 \\ \bottomrule
\end{tabular}
\caption{The Attack Success Rate Difference (ASRD) of SynAtk on three benchmarks.}
\label{tab5}

\end{table}
\end{document}